\documentclass[10pt,twocolumn,letterpaper]{article}

\usepackage[pagenumbers]{cvpr} 

\usepackage{graphicx}
\usepackage{amsmath}
\usepackage{amssymb}
\usepackage{booktabs}
\usepackage[accsupp]{axessibility}  

\usepackage[pagebackref,breaklinks,colorlinks]{hyperref}

\usepackage[capitalize]{cleveref}
\crefname{section}{Sec.}{Secs.}
\Crefname{section}{Section}{Sections}
\Crefname{table}{Table}{Tables}
\crefname{table}{Tab.}{Tabs.}

\def\cvprPaperID{8137} 
\def\confName{CVPR}
\def\confYear{2023}

\begin{document}

\title{Egocentric Auditory Attention Localization in Conversations}

\author{Fiona Ryan$^{1,2*}$, Hao Jiang$^{2}$, Abhinav Shukla$^{2}$, James M. Rehg$^{1,2}$, Vamsi Krishna Ithapu$^{2}$ \\
Georgia Institute of Technology$^{1}$, Meta Reality Labs Research$^{2}$ \\
{\tt\small \{fkryan, rehg\}@gatech.edu, \{haojiang, ithapu\}@meta.com, abhinav.shukla.research@gmail.com}}
\maketitle

{\let\thefootnote\relax\footnote{{*Work done during internship at Meta.}}}

\begin{abstract}
   In a noisy conversation environment such as a dinner party, people often exhibit selective auditory attention, or the ability to focus on a particular speaker while tuning out others. Recognizing who somebody is listening to in a conversation is essential for developing technologies that can understand social behavior and devices that can augment human hearing by amplifying particular sound sources. The computer vision and audio research communities have made great strides towards recognizing sound sources and speakers in scenes. In this work, we take a step further by focusing on the problem of localizing auditory attention targets in egocentric video, or detecting who in a camera wearer's field of view they are listening to. To tackle the new and challenging Selective Auditory Attention Localization problem, we propose an end-to-end deep learning approach that uses egocentric video and multichannel audio to predict the heatmap of the camera wearer's auditory attention. Our approach leverages spatiotemporal audiovisual features and holistic reasoning about the scene to make predictions, and outperforms a set of baselines on a challenging multi-speaker conversation dataset. Project page: \url{https://fkryan.github.io/saal}
   
\end{abstract}


\section{Introduction} \label{sec:intro}

\begin{figure}[t]
  \centering
   \includegraphics[width=1.0\linewidth]{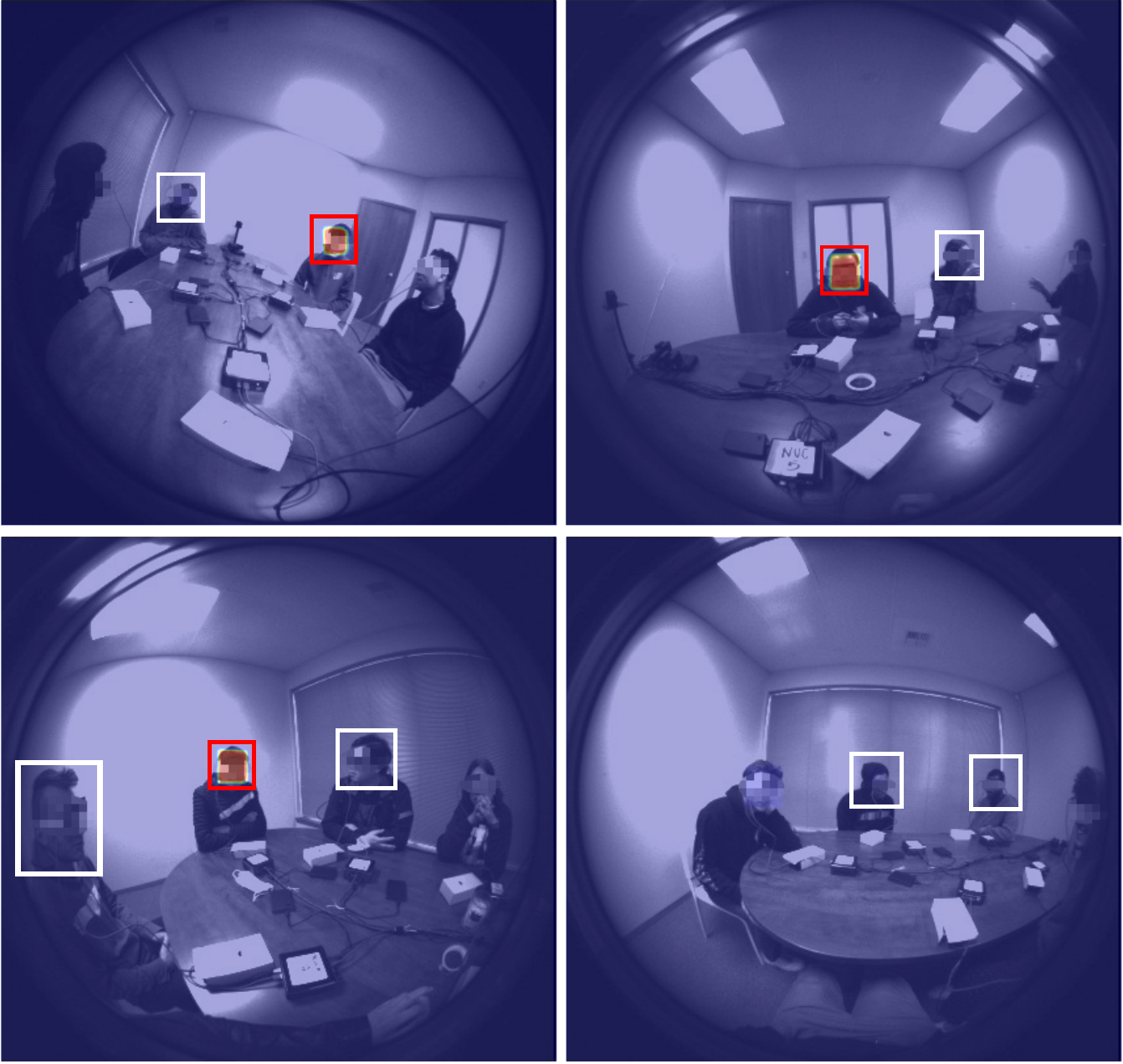}
   \caption{We address the novel task of Selective Auditory Attention Localization in multi-speaker environments: given egocentric video and multichannel audio, we predict which people, if any, the camera wearer is listening to. We show our model's predicted heatmaps, with red bounding boxes denoting ground truth auditory attention targets and white boxes denoting non-attended speakers.}
   \vspace{-1em}
   \label{fig:onecol}
\end{figure}

One of the primary goals of wearable computing devices like augmented reality (AR) glasses is to enhance human perceptual and cognitive capabilities. This includes helping people have natural conversations in settings with high noise level (e.g. coffee shops, restaurants, etc.) by selectively amplifying certain speakers while suppressing noise and the voices of background speakers.
This desired effect mirrors selective auditory attention (SAA), or humans' ability to intentionally focus on certain sounds while tuning out others. People exercise SAA 
in everyday conversational settings; at restaurants people tune out the voices at adjacent tables, and in group social settings, such as dining at a large table or socializing at a party, people often engage in conversations with smaller subsets of people while others converse in close proximity. Being able to determine which speaker(s) a person is selectively listening to is important for developing systems that can aid communication in noisy environments and assist  people with hearing loss. 

While the computer vision community has made strides towards understanding conversational dynamics with large-scale datasets like Ego4D \cite{Ego4D}, AVA-ActiveSpeaker \cite{AVAActiveSpeaker}, VoxConverse\cite{chung2020spot}, and AMI \cite{kraaij2005ami}, the problem of modeling selective listening behavior has not yet been addressed. In fact, determining auditory attention among competing sound signals has traditionally been approached using neurophysiological sensing \cite{alickovic2019tutorial, kaya2017modelling}. These approaches involve a controlled listening task, where competing audio signals are played simultaneously and a listener selectively listens to one. Statistical models are then used to correlate brain activity and the attended sound signal. However, the sensing approach is obtrusive and  currently not feasible for use in realistic conversational settings in daily life.

In this work, we approach modeling SAA from an egocentric audiovisual perspective. The egocentric
setting provides a compelling lens to study social conversational dynamics, as it captures both the audio and visual stimuli present in a scene and how the camera wearer orients their view in response to them. We hypothesize that the behavior of the wearer that is implicitly captured in the egocentric video and multichannel audio can facilitate the prediction of auditory attention targets. 
To this end, we introduce and formulate the novel task of \textbf{Selective Auditory Attention Localization (SAAL)}, which uses egocentric video and multichannel audio to localize the target of auditory attention in egocentric video. We specifically target multi-speaker conversation scenarios where the camera wearer must selectively attend to certain speaker(s) while tuning out others. This challenging setting is representative of everyday noisy conversation environments and highlights the complexity of modeling auditory attention in naturalistic settings.

We propose a deep audiovisual video network for SAAL that leverages both appearance-based features from the egocentric video stream and spatial audio information from multichannel audio. Our key
insight is to extract a spatiotemporal feature representation from each modality and use a transformer on the fused features to reason holistically about the scene. Our contributions are:
\vspace{-0.5em}

\begin{itemize}
\itemsep0em
  \item We introduce the novel problem of Selective Auditory Attention Localization (SAAL) as an egocentric multi-modal learning task. 
  \item We propose a new architecture for SAAL. Our model extracts spatiotemporal video and audio features and uses a transformer to refine selection of an attention target by reasoning globally about the scene.
  \item We evaluate our approach on a challenging multi-speaker conversation dataset and demonstrate our model's superiority over intuitive baselines and approaches based on Active Speaker Localization (ASL).
  \item We conduct thorough experiments to give insight into our multimodal architecture design, effective multichannel audio and visual input representations, and the relationship between SAAL and ASL.
\end{itemize}
\vspace{-0.5em}

\begin{figure*}[t]
  \centering
  \includegraphics[width=0.97\linewidth]{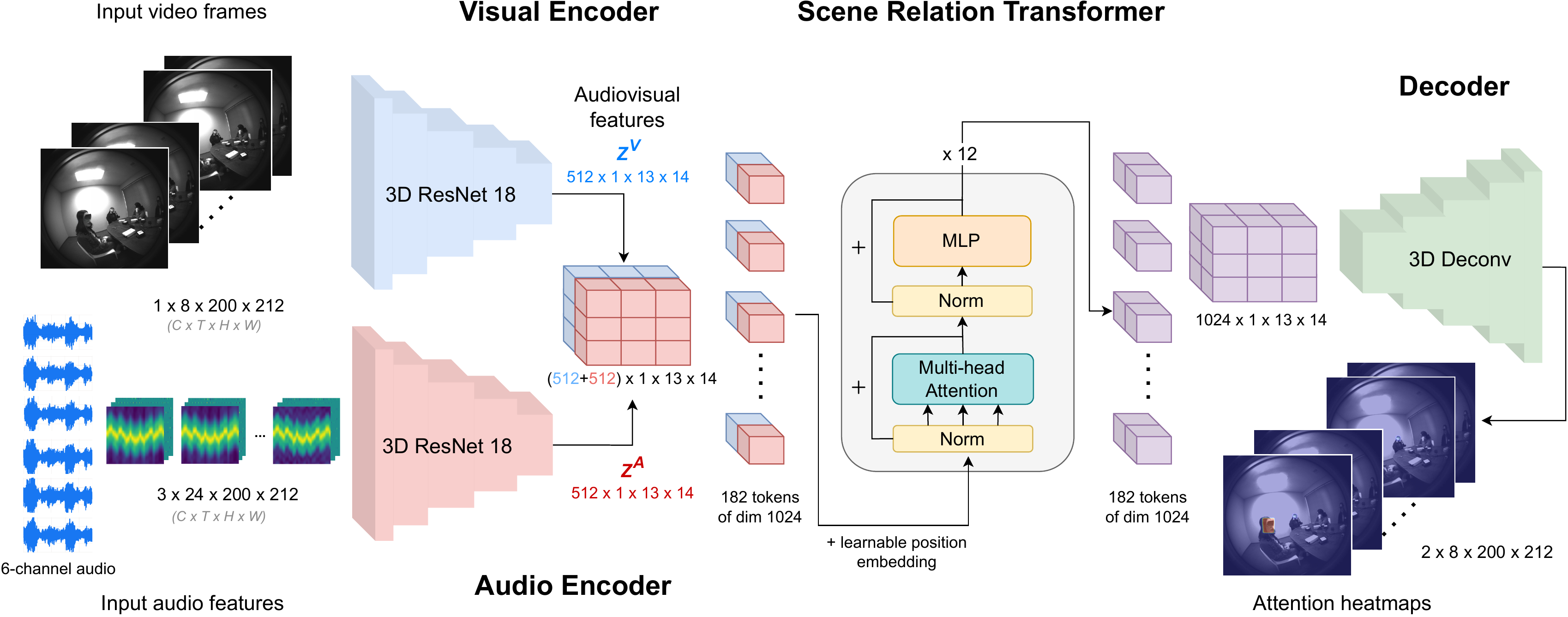}
  \caption{Our architecture for predicting auditory attention heatmaps from egocentric video and multichannel audio: the \textbf{Video Encoder} and \textbf{Audio Encoder} embed spatiotemporal features from the video and spatial audio modalities respectively. The audio and visual features are fused and passed to the \textbf{Scene Relation Transformer}, which models the relationships between different regions of the scene. Finally, the features are passed to the convolutional \textbf{Decoder}, which produces the predicted attention heatmaps for the input video frames.}
  \vspace{-0.5em}
  \label{fig:architecture}
\end{figure*}


\section{Related Work} \label{sec:related}

\noindent\textbf{Computational Models for Auditory Attention} 
Attention is typically categorized into two types:
bottom-up, where a stimulus is likely to cause attention due to inherent salient properties, 
and top-down, where a person intentionally focuses on a certain stimulus. 
Top-down, or selective auditory attention has been studied by correlating the attended sound signal with the neural response measured by EEG / MEG using statistical models \cite{ding2012emergence,mesgarani2012selective, mirkovic2015decoding, o2015attentional, o2017neural, Bleichner2016, Horton2014, haghighi2018eeg, Bednar2020, Biesmans2016,fuglsang2017noise, van2016eeg, akram2016robust}, and more recently deep neural networks \cite{Kuruvila2021ExtractingTA, Geravanchizadeh2021}. This is done in a controlled listening scenario where competing sounds, such as audio books, are played at once and a participant is instructed to listen to a particular sound. Models are used to predict which sound source is attended and/or decode the attended signal. The sources may be combined into a single channel or played in separate ears, as in the dichotic listening task paradigm \cite{broadbent1956successive}. Some works have used acoustic simulation methods to reflect more realistic multi-speaker environments \cite{fuglsang2017noise, van2016eeg, Bednar2020}. However, these studies have not addressed modeling SAA in in-the-wild social conversation scenarios. Beyond using brain activity to predict auditory attention, Lu et al. develop a head movement-based approach for predicting the attended speaker in a conversation \cite{LuLSTM}. However, this work focuses on single group conversations where there is typically only one speaker.

A smaller body of work focuses on modeling bottom-up auditory attention, or the qualities of a sound signal that make it likely to be attended to. Works in this domain construct representations of audio signals and evaluate their ability to reflect human judgment of saliency \cite{kayser2005mechanisms, duangudom2007using, kaya2014investigating, kim2014automatic, tordini2013toward, tordini2015loud, kothinti2021auditory}, distinguish between changes in perceptual qualities of sound \cite{kaya2012temporal}, or perform on a downstream speech recognition task \cite{kalinli2007saliency}. In this work, we focus on top-down, or selective auditory attention.

\noindent\textbf{Active Speaker Detection \& Localization} 
Our task builds upon a rich foundation of work on Active Speaker Detection (ASD) and Active Speaker Localization (ASL). Recognizing speech activity is an established problem in computer vision, with early approaches correlating video and audio streams \cite{cutler2000look}, or using purely visual features \cite{saenko2005visual, everingham2009taking, haider2012towards} to determine if a person is speaking. The AVA-ActiveSpeaker dataset \cite{AVAActiveSpeaker} has accelerated deep learning approaches for ASD by providing a large-scale dataset and a benchmark.
Methods use deep learning to extract visual features from a head bounding box track and combine them with features from monoaural scene audio to classify the person as speaking and audible, speaking and inaudible, or not speaking in each frame 
\cite{Tao2021SomeoneSpeaking, Zhang2019MultiTaskLF, kopuklu2021design, Alcazar_2020_CVPR, AlcazarMaas, Min2022LearningLS, Alcazar2022EndtoEndAS, ASDTransformer}. 
Some approaches additionally explicitly model the relationships between multiple people within a scene to refine classification of speakers \cite{kopuklu2021design, Alcazar_2020_CVPR, Min2022LearningLS}. Xiong et al. \cite{Xiong2022LookListenMC} develop an architecture to jointly perform ASD and speech enhancement.

Other methods approach the task of ASL, which seeks to 
localize speakers spatially within the scene rather than classifying bounding box tracks  \cite{GebruClustering, chakravarty2015s, wissing2021data, pu2020active, ban2019variational, gatica2007audiovisual, zhang2008boosting, alameda2011finding}. Several use multichannel audio to incorporate directional audio information  \cite{GebruClustering, chakravarty2015s, wissing2021data, ban2019variational, gatica2007audiovisual, zhang2008boosting, alameda2011finding}.
Recently, Jiang et al. \cite{jiang2022egocentricdeep} expanded upon these approaches by developing a deep model that also leverages multichannel audio to localize speakers in the challenging egocentric setting, which contains motion blur and rapid changes of field of view (FOV), using the EasyCom dataset \cite{EasyCom}. 
Our problem setting is different from ASD and ASL in that we seek to determine which of the speakers the camera wearer is \emph{listening to} using egocentric audiovisual cues. 
Importantly, active speakers are not necessarily auditorily attended; 
they can be background people or nearby people engaged in a different conversation. Thus, SAAL demands a novel approach.

\noindent\textbf{Modeling of Conversation Dynamics and Social Interactions}
SAAL is also related to work on modeling social interactions and conversational dynamics. 
Prior works have explored detecting social interactions in egocentric video \cite{aghaei2016whom, fathi2012social, bano2018multimodal}. The recent Ego4D dataset and benchmark suite \cite{Ego4D} introduces the ``Talking to Me" task for identifying which people are speaking to the camera wearer in egocentric video. This task is closely related to SAAL in that it models dyadic conversational behavior between the camera wearer and people in their FOV. However our task differs by modeling egocentric listening behavior, particularly in the presence of multiple speakers. Researchers have also explored modeling conversational turn taking behaviors \cite{Truong2021RightToTalk} as well as predicting listener motion in dyadic conversations \cite{ng2022learning}. To our knowledge, we are the first to addressthe problem of modeling auditory attention using egocentric video, wearable audio, or both modalities together.

\noindent\textbf{Egocentric Visual Attention Prediction}
Our problem also relates to egocentric visual attention prediction, 
which regresses a heatmap localizing the target of visual attention in egocentric video.
Approaches for this task use eye tracking data as ground truth for attention targets 
and develop models to predict the attention location from video.
Researchers have proposed several vision models for this task \cite{AlNaser2019OGazeGP, Huang2018PredictingGI, LiEgocentric2013, Tavakoli2019DiggingDI, Lai2022}, 
including some that jointly model gaze and action \cite{HuangMutual, LiBeholder2021} 
or gaze and IMU \cite{Thakur2021}. 
Importantly, visual attention and auditory attention are not the same; people will generally not always look at the person they are conversing with. Rather, SAAL can be considered as an auditory variant of egocentric visual attention prediction, and SAAL additionally demands the use of audio cues.

\noindent\textbf{Audiovisual Representation Learning}
Our work relates to a larger body of research on learning effective audiovisual feature representations \cite{Jansen2020CoincidenceCA, owens2016ambient, owens2018audio, akbari2021vatt, alayrac2020self, korbar2018cooperative, patrick2020multi, alwassel2020self, nagrani2021attention} and localizing sound sources in video \cite{hu2019deep, hu2021class, liu2022visual, qian2020multiple, senocak2018learning, afouras2020self, chen2021localizing, senocak2022learning, Arandjelovic_2018_ECCV, mo2022localizing}. These methods learn semantic correspondences between visual and audio features, typically using contrastive training paradigms. In contrast to these approaches, we leverage multichannel audio and employ a fusion mechanism that enforces spatial alignment between the visual and spatial audio domains, allowing us to reason globally about the audiovisual content in different regions of the scene.


\section{Method} \label{sec:method}

\subsection{Problem Definition}
At each time $t$, we predict an auditory attention heatmap $\mathcal{H}_t$,
given the raw audiovisual input $\{I_n, a_n | n = 0 ... t\}$, where
$I_n$ is an egocentric video frame at time $n$ and $a_n$ is a multi-channel audio segment aligned with the video frame $I_n$.
Each pixel $\mathcal{H}_{t}(i,j)$ has a value in (0,1) and indicates the probability of the camera wearer's auditory attention. Video ${I}$ captures the movement of both the camera wearer and the other people in the scene, as well as information about their head and body poses, facial expression,  mouth movement, and overall scene layout. Multichannel audio $a$, from a microphone array, captures people's speech as well as other spatialized sound sources. We use video from a 180-degree fish-eye camera with resolution $200\times212$ and audio from a 6-channel microphone array. In this paper, we demonstrate how to use $\{I,a\}$ to infer $\mathcal{H}$, where we construct $\mathcal{H}\textsubscript{label}$ by setting all pixels within the head bounding box of an attended speaker in a frame as 1.

\subsection{Model} \label{subsec:model}
SAAL is a complex task that involves understanding both the camera wearer's behavior and the activity of the people in the scene.
We design a deep multimodal model for this task to have 4 key abilities:
\textbf{(1)} It extracts relevant visual features such as lip movement, the relative positions and orientations of heads, and social body language like when a person looks at the camera wearer. \textbf{(2)} It leverages multichannel audio to construct a spatial representation of voice activity occurring around the camera wearer. \textbf{(3)} It encodes temporal information like how the wearer moves their head in response to stimuli as well as the movement of people in the scene and pattern of voice activity. \textbf{(4)} It reasons globally about the audiovisual content of scene as a whole to select an attention target; while an active speaker detector identifies all speakers in a scene, determining attention demands holistic reasoning about different potential attention targets in a scene to select the most likely one(s). 

To this end, our method consists of 4 main components as shown in Figure \ref{fig:architecture}: the \textit{Visual Encoder} and \textit{Audio Encoder} to extract spatiotemporal features from the video and multichannel audio inputs respectively, the \textit{Scene Relation Transformer} to refine attention localization by modeling the relationships between audiovisual features in different spatial regions of the scene, and a convolutional \textit{Decoder} to produce an auditory attention heatmap for each input frame.

\noindent\textbf{Visual Encoder}
The Visual Encoder $\mathcal{V}$ takes the visual input $X^{V}$ as a stack of $200 \times 212$ frames over a time interval $T$ and extracts appearance-based features about the scene. We construct $X^{V}$ by cropping the people's heads in the raw grayscale video frames and filling the background with black. This representation leverages 1) our focus on conversational attention, and 2) the effectiveness of face detection methods. The input is a 3D video tensor of head tubes that represents not only how the people in the scene move, how they are laid out in space, where they look, their facial expression, mouth movement, and who they talk to, but also the head movement of the wearer over time. In our experiments, we compare this representation to using the raw image as well as a binary map of the bounding boxes. We find that this representation improves performance by removing background distractions while retaining appearance-based facial features. Additionally, this representation makes the model more generalizable to different scenes. We use a 3D ResNet-18 network \cite{tran2018closer} as our backbone, which uses 3D convolutions to capture both spatial and temporal features. The result of $\mathcal{V}(X^V)$ is a spatiotemporal visual feature map $\mathcal{Z}^V$ of dimension $512\times1\times13\times14$, where 512 is the channel dimension, 1 is the temporal dimension, and $13$ and $14$ correspond to the height and width. We use 8 frames as input with temporal stride 3, so each input $X^V$ spans a 24 frame window, which is 0.8 seconds in 30fps video.

\noindent\textbf{Audio Encoder} The Audio Encoder $\mathcal{A}$ extracts spatial audio features from the multichannel audio in the input clip to encode information about the conversational speech patterns and spatially localize speakers. We process the audio into 2000 sample $\times$ 6 channel frames corresponding to each frame in the video input clip. We construct input features $X^{A}$ by calculating the complex spectogram with 201 frequency bins, window length 20, and hop length 10 and stacking these vertically for the 6 channels. We found this performs slightly better than concatenating along the channel dimension. We additionally construct a feature representation of the cross correlation between all pairs of channels as used in Jiang et al. \cite{jiang2022egocentricdeep} and stack the real and complex parts of the spectogram and the channel correlation features along the channel dimension. We then resize the audio features to $200 \times 212$ spatially in order to extract feature maps that align with the visual domain. Due to the importance of capturing dense temporal audio features to predict auditory attention, such as the pattern of speech, we pass all 24 audio frames in the clip to $\mathcal{A}$, which uses another 3D ResNet-18 backbone. The resultant feature map $\mathcal{Z}^A$ is of size $512 \times 3 \times 13 \times 14$. We average pool the temporal dimension to 1 to match the dimension of $\mathcal{Z}^V$.

\noindent\textbf{Scene Relation Transformer} 
From $\mathcal{A}$ and $\mathcal{V}$, we extract visual and spatial audio features maps $\mathcal{Z}^{V}$ and $\mathcal{Z}^{A}$, which we concatenate along the channel dimension to form $\mathcal{Z}^{AV} \in 1024 \times 1 \times 13 \times 14$. We choose to concatenate the features along the channel dimension because our approach importantly leverages multichannel audio, which can extract a spatial representation of voice activity in the scene so that the visual and audio domains can be spatially aligned. 

\begin{table*}
\small
  \centering
  \begin{tabular}{@{}lrrrr@{}}
    \toprule
    Split & Total frames & Frames with 1+ attention target & Avg. heads per frame & Avg. speakers per frame \\
    \midrule
    Train & 1,416,262 & 881,931 & 3.05 & 1.43 \\
    Test & 571,027 & 367,536 & 3.06 & 1.45 \\
    \bottomrule
  \end{tabular}
  \caption{Dataset statistics: our evaluation dataset of multi-speaker conversations is representative of challenging conversation environments where several people and speakers are present. Note that we use $\frac{1}{3}$ of these frames for training and evaluation with temporal stride 3.}
  \vspace{-0.5em}

  \label{tab:dataset}
\end{table*}

We use a transformer $\mathcal{T}$ to further refine this fused multimodal feature map by modeling correlations between the features in different regions of the scene to holistically select a likely attention target.
As in the transformer literature \cite{Vit, arnab2021vivit}, we treat each spatial location $Z^{AV}_{1,h,w}$ as a spatiotemporal patch of embedded dimension $D=512$, and flatten the feature map into a sequence of length $L=1 \times 13 \times 14 = 182$. A learnable position embedding $E \in \mathbb{R}^{L\times D}$ is added to preserve absolute and relative positional information, which is particularly important for the egocentric setting where the camera wearer orients their view of the scene in response to stimuli. The transformer consists of 12 transformer blocks \cite{vaswani2017attention} where each computes self-attention on the audiovisual feature map as
$ \textit{Softmax}(QK^T/\sqrt{D}) V $,
where $Q$, $K$, and $V$ are learned linear projections on the multimodal spatiotemporal patches. In this way, the model learns correlations between audiovisual features occurring in different regions of the scene and can suppress or enhance values accordingly to refine attention prediction.

\noindent\textbf{Decoder} A decoder consisting of 4 3D tranpose convolutions and 1 final convolutional classifier layer upsamples the resulting feature map $\mathcal{T}(\mathcal{Z}^{AV})$ to produce the attention class heatmap $\mathcal{H}$\textsubscript{pred} of size $2 \times T \times H \times W$. The 2 channels reflect the not attended and attended classes.

\subsection{Implementation \& Model Training}
We train end-to-end with pixel-wise cross entropy loss. We initialize $\mathcal{V}$ and $\mathcal{A}$ with pretrained weights on Kinetics 400 \cite{carreira2017quo}. The transformer consists of 12 transformer blocks using multi-head self-attention with 8 heads \cite{vaswani2017attention}. We use the Adam optimizer \cite{kingma2014adam} and learning rate 1e-4. The training procedure converges quickly in less than 10 epochs.


\section{Experimental Results} \label{sec:experiments}

\begin{figure}[]
  \centering
   \includegraphics[width=0.85\linewidth]{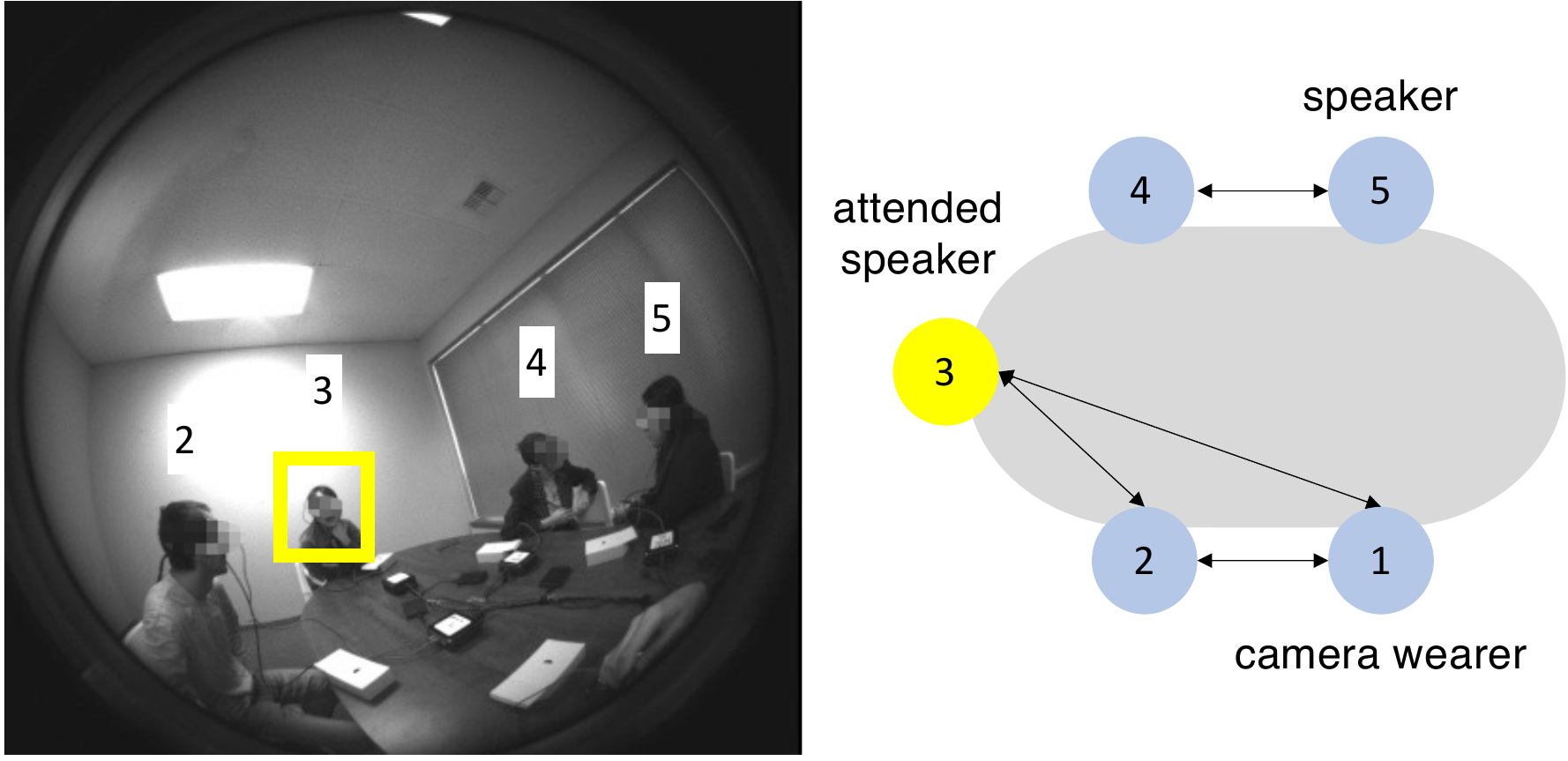}

   \caption{In our multi-speaker evaluation dataset, participants are divided into 2 subgroups that converse simultaneously. A person is labeled as being an auditory attention target if they are both speaking and within the camera wearer's conversation group.}
   \vspace{-1em}
   \label{fig:dataset}
\end{figure}

\subsection{Evaluation Dataset \& Criteria}
To our knowledge there is no existing dataset with associated labels suited to our task. 
While egocentric AV datasets like Ego4D\cite{Ego4D} and EasyCom\cite{EasyCom} capture egocentric conversations, they have few situations with multiple people speaking at the same time, with the result that SAAL reduces to ASL. To evaluate our approach, we collected a dataset of complex multi-speaker conversations, where groups converse in close proximity. The dataset consists of 50 participants conversing in groups of 5, where each person wears a headset with a camera and microphone array. In total, there are $\sim$20 hours of egocentric video. 

In each 10-minute recording, a group of 5 people sits at a table in a meeting room and is partitioned into two conversational groups. Participants are instructed to only listen to and engage in conversations with their own group. In this way, we design a naturalistic social version of traditional dichotic listening tasks, where participants intentionally focus on certain speakers while tuning out others. By splitting  the group into smaller  groups, we simulate conversational layouts that may occur at a large dinner table, coffee shop, restaurant, or party, where groups of people converse in close proximity with others. As illustrated in Figure \ref{fig:dataset}, we construct ground truth auditory attention labels by determining that the camera wearer is listening to a given person if they are both speaking and in the wearer's conversation group. The dataset was collected under an approved IRB.

Each participant wears a headset that includes an Intel SLAM camera and a 6-microphone array. 
We record the fish-eye grayscale video and the audio from the 6 microphones simultaneously for all the wearers in each session. We generate speaker activity and head bounding box annotations automatically by using person tracking and wearer speech detection algorithms, and leveraging synchronization between each person's egocentric audio and video in each session. We obtain active speaker labels for each person by using their own headset's microphone array to determine if they are speaking, using a deep model adapted from \cite{jiang2022egocentricdeep} trained on synthetic data for the headset's microphone array configuration. We then use a 3D tracking-by-detection method that uses camera pose from the SLAM camera to annotate head bounding boxes and identities for each person in each egocentric FOV. From these person tracking results and speaker activity labels for each person, we have a set of visible head bounding boxes for each egocentric frame along with a label for each box indicating if that person is speaking. We then use the identity tracking and known conversation group assignments to label if each person is auditorily attended to by the camera wearer.

Due to the wide 180-degree FOV of the camera, the auditorily attended person is almost always in the FOV, so we constrain our modeling approach to this scenario. Additionally, by design of the dataset, there are typically several people and often multiple speakers within FOV, as shown in Table \ref{tab:dataset}. Along with the inherent challenges of motion blur and rapid head movements in the egocentric modality, this makes our dataset a rich means for studying auditory attention in complex and naturalistic conversation environments.

We evaluate the result of auditory attention localization using mean average precision (mAP) on the task of classifying each person in the FOV as being a target of auditory attention or not. From the output heatmap $\mathcal{H}$\textsubscript{pred}, we calculate the auditory attention score for each person by taking the maximum value of $\textit{Softmax}(\mathcal{H}\textsubscript{pred})$ in channel 1 (the attended class) in the region of their head bounding box. 
We split our data and use 70\% data for training and 30\% for testing, with no overlapping participants between splits.

\subsection{Competing methods}
Because SAAL is a new problem, there are no previous methods that can be used directly to solve this problem. We therefore compare our approach to several naive baseline methods as well as a set of baselines that adapt Jiang et al.'s multichannel audiovisual ASL architecture (MAVASL) \cite{jiang2022egocentricdeep} for our task.
The methods we compare are the following:

\textbf{(1)} Our proposed method with different audio and visual input representations:
In our experiments, we label variations of our method as ``Ours - [visual] \& [audio]", where [visual] is the visual input representation type and [audio] is the audio input representation type.
The visual input representations we consider are \textbf{Image} (the raw image), 
\textbf{Bbox} (a binary map of the bounding boxes), 
and \textbf{Heads} (the cropped heads from the raw image on a black background). 
The audio representations are \textbf{Channel Corr} (the channel correlation features as described in Section \ref{subsec:model}), \textbf{Channel Corr + Spectogram} (the channel correlation features concatenated with the real and complex parts of the multichannel spectogram), and 
\textbf{ASL} (active speaker localization output maps from MAVASL). 
We report two different training strategies for using MAVASL to generate ASL maps for our dataset: \textbf{ASL\textsubscript{synthetic}}, which trains the ASL model on synthetic data created from VTCK \cite{VTCKveaux2017cstr} and EasyCom, 
and \textbf{ASL\textsubscript{real}}, which directly trains the ASL model on our dataset using the active speaker labels. In this way, we compare using audio features constructed from the raw multichannel audio with using ASL maps generated from an existing model, both with and without the advantage of tuning to our dataset.

\textbf{(2)} Naive baselines using ground truth ASL labels and heuristic strategies: We first report the SAAL mAP from simply selecting all active speakers as being attended using the ground truth ASL labels (Perfect ASL). We also report 2 baselines that use the center prior, which is the intuition that the attended person is likely near the center of the FOV because people often face the speaker they are paying attention to. Similar center prior baselines are used in the egocentric visual attention literature. We construct 2 center prior baselines: CP--I finds the head bounding box nearest to the center of FOV and selects it as the attention target if they are speaking, and otherwise predicts that there is no attention in the frame. CP--II selects the speaking person closest to the center of the FOV as the target of attention. We also construct 2 baselines that select the ``closest" speaker based on the size of their bounding box. We can reason that the person with the largest head bounding box area is likely the closest to the camera wearer, and if speaking, likely appears the loudest. We consider 2 baselines based on this strategy: CS--I selects the ``closest" person only if they are speaking, and CS--II selects the ``closest" speaker. We note that because these methods depend on the dataset's ground truth ASL labels, they are not realistic results, and are not a fair comparison to our method. However, we report them to demonstrate that Perfect ASL and heuristic strategies perform poorly on our evaluation dataset. This clearly implies that SAAL is a complicated task that demands a novel approach, and that our evaluation dataset is rich enough to demonstrate the complexity of this task.

\textbf{(4)} Networks adapted from MAVASL: 
The first adaption (MAVASL-I) is to directly use MAVASL to localize active speakers and use this as the auditory attention heatmap. The second adaptation (MAVASL-II) directly trains the network for the task of SAAL instead of ASL. The third adaptation (MAVASL-III) is similar to (MAVASL-II) but uses a multi-frame video input (3-frame) to the AV network to further incorporate temporal information. MAVASL consists of an Audio Network, which predicts a coarse spatial voice activity map from multichannel audio, and an AV Network, which uses this voice activity map and the image to predict a final ASL map. For MAVASL-II and III, we achieve best results by supervising the audio network in MAVASL using the ASL labels from our dataset and the AV network with the auditory attention labels. We use \textbf{Heads} as the visual input representation. Because MAVASL is a frame-based network, we additionally apply temporal smoothing with a window size of 10 frames as in the original method.

\begin{table}
\small
  \centering
  \begin{tabular}{@{}lc@{}}
    \toprule
    Method & mAP (\%) \\
    \midrule  
    Perfect ASL*      & 47.99 \\
    CP--I*  & 63.55  \\
    CP--II* & 51.48  \\
    CS--I* & 53.86 \\
    CS--II* & 49.47 \\
    \hline
    MAVASL--I         & 59.11  \\
    MAVASL--II        & 75.20  \\
    MAVASL--III       & 72.90   \\
    \hline
    Ours--Bbox \& ASL\textsubscript{synthetic} & 75.93 \\
    Ours--Bbox \& ASL\textsubscript{real} & 74.97 \\
    Ours--Bbox \& Channel Corr & 80.41 \\
    Ours--Bbox \& Channel Corr + Spectogram & 80.31 \\
    Ours--Image \& ASL\textsubscript{synthetic} & 72.20 \\
    Ours--Image \& ASL\textsubscript{real} & 70.04 \\
    Ours--Image \& Channel Corr & 76.52 \\
    Ours--Image \& Channel Corr + Spectogram & 76.95 \\
    Ours--Heads \& ASL\textsubscript{synthetic} & 76.72 \\
    Ours--Heads \& ASL\textsubscript{real} & 77.11 \\
    Ours--Heads \& Channel Corr & 82.35 \\
    \textbf{Ours--Heads \& Channel Corr + Spectogram} & \textbf{82.94} \\
    \bottomrule
  \end{tabular}
  \caption{Comparison results on the multi-speaker conversation dataset. (*) denotes methods that use ground truth ASL, which is not given to our model.}
  \label{tab:example}
\end{table}

\subsection{Comparison results}

\begin{figure*}
  \centering
  \includegraphics[width=0.84\linewidth]{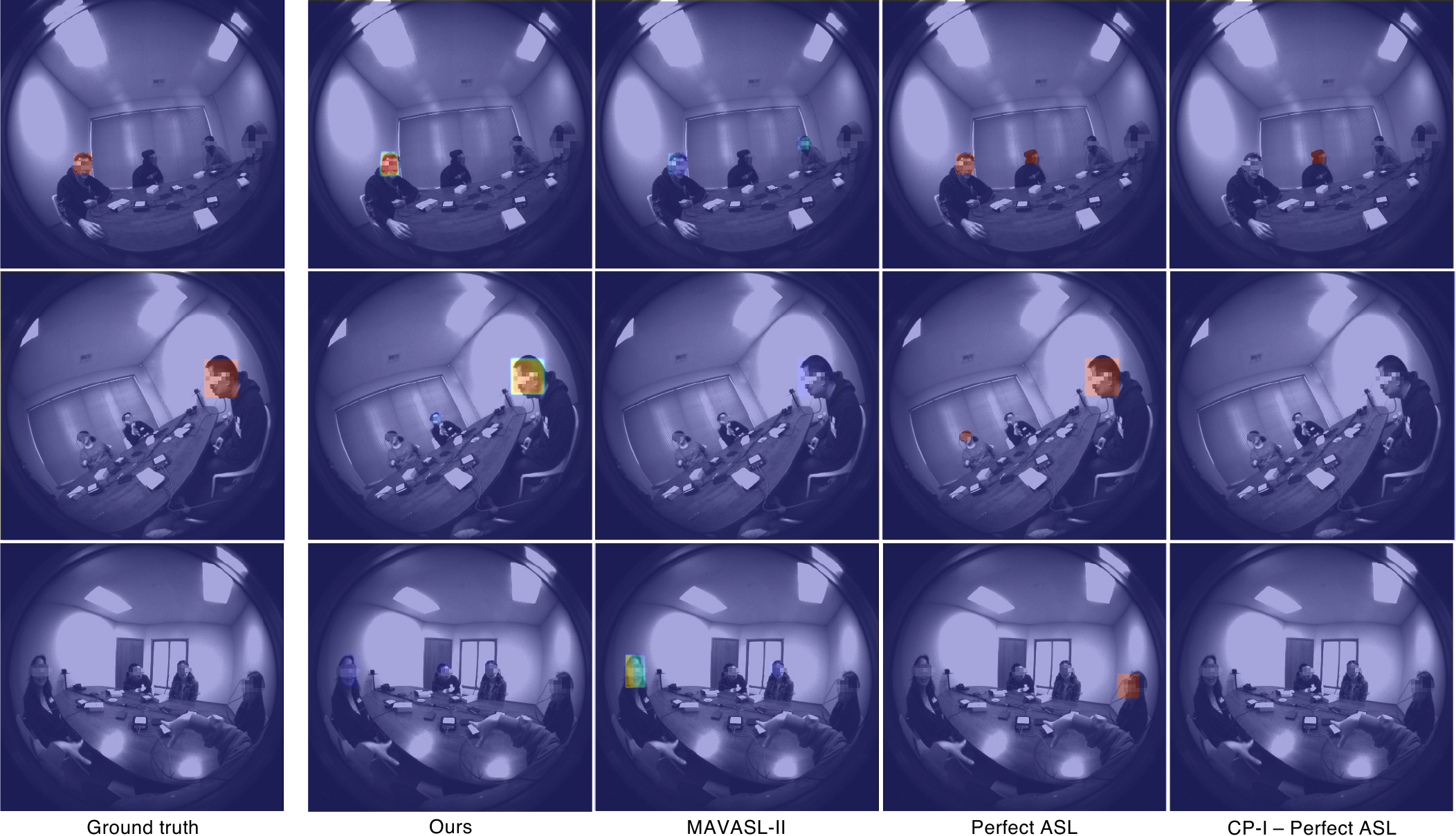}
  \caption{We compare our output heatmaps to competing methods on challenging cases with multiple speakers, the attended speaker being far from the center, or no auditory attention. Our method performs well in these scenarios and more consistently identifies who is and isn't attended than MAVASL-II. Perfect ASL is prone to false positives by selecting all speakers and CPI is prone to false negatives by only considering the person closest to the center, demonstrating that ASL is not enough to solve SAAL. (\textbf{Best viewed in color}).}
  \vspace{-1em}
  \label{fig:challenging_viz}
\end{figure*}

As shown in Table \ref{tab:example}, our proposed method outperforms the baselines on mAP, 
including baselines that leverage ground truth active speaker labels.
We observe that the Perfect ASL and naive baselines perform poorly on our dataset, 
illustrating that reducing the problem of SAAL to ASL is not viable in complex and challenging conversation environments with multiple speakers. 
The highest mAP achieved from adapting Jiang et al.'s ASL architecture is MAVASL-II, which obtains 75.20\% mAP. However, our best approach achieves a gain of 7.74\% mAP over this baseline, 
demonstrating that the task of SAAL demands a novel architecture design. We qualitatively compare our results against these baselines in Figure \ref{fig:challenging_viz}.

From our experiments investigating different combinations of audio and visual input representations, 
we observe that the combination of \textbf{Heads} and \textbf{Channel Corr + Spectogram} produces the best results. 
In fact, we observe that \textbf{Bbox}, the binary bounding box location map, performs better as an input representation than the raw image. We can interpret this result as showing the importance of relative position and sizes of heads to the problem of SAAL. \textbf{Channel Corr} and \textbf{Channel Corr + Spectogram}, which contain information from the raw multichannel audio signal, 
generally perform better as the spatial audio input representation than the pre-computed ASL maps, both real and synthetically trained. 
We hypothesize that this is because the role of audio features in determining auditory attention is greater than just localizing active speakers; SAAL additionally involves correlating different speech signals with the camera wearer's egocentric movement, using the wearer's own voice activity as an indicator of conversational turn taking behavior (e.g. whether they are speaking or listening), and reasoning about the relative volume of different speech signals. Notably, the pre-computed ASL maps have the advantage of being spatially aligned with the visual modality, which the spectograms and channel correlation features do not have. The performance gains from using \textbf{Channel Corr} and \textbf{Channel Corr + Spectogram} indicate that our Audio Encoder is able to learn this alignment and construct a spatial representation of audio activity from these features.

\begin{table}
\small
  \centering
  \begin{tabular}{@{}lc@{}}
    \toprule
    Method & mAP (\%) \\
    \midrule
    Visual only & 56.81 \\
    Visual only + Transformer & 57.70 \\
    Audio only & 76.30 \\
    Audio only + Transformer & 75.84 \\
    Audiovisual & 80.10 \\
    Audiovisual\textsubscript{1-channel} + Transformer & 71.47 \\
    Audiovisual\textsubscript{2-channel} + Transformer & 76.73 \\
    Audiovisual\textsubscript{4-channel} + Transformer & 80.99 \\
    \textbf{Audiovisual + Transformer} & \textbf{82.94} \\
    \bottomrule
  \end{tabular}
  \caption{Model ablation: All studies use Heads as the visual input and Channel Corr + Spectogram as the audio input.}
  \label{tab:ablation}
\end{table}

\subsection{Analysis}
\noindent \textbf{Model ablation} We further investigate the contribution of each component of our best model to its overall performance in Table \ref{tab:ablation}. 
We observe that the visual modality alone is much weaker than the spatial audio modality, 
and that clearly audio plays an important role. We also ablate the number of audio channels used, observing that removing the spatial component of the audio (Audiovisual\textsubscript{1-channel} + Transformer) significantly reduces performance of our overall model. We see that spatial audio is an important cue for this task, but using slightly less channels can still achieve effective performance. Adding the transformer to the audiovisual features results in a gain of 2.84 mAP, demonstrating that the transformer improves performance by refining the embedded audiovisual feature maps through modeling relationships across spatial regions of the scene.

\begin{table}
\small
  \centering
  \begin{tabular}{@{}lc@{}}
    \toprule
    Method & mAP (\%) \\
    \midrule
    Audiovisual + Transformer (no spatial alignment) & 80.57 \\
    \textbf{Audiovisual + Transformer (spatial alignment)} & \textbf{82.94} \\
    \bottomrule
  \end{tabular}
  \caption{Spatial modality alignment in the transformer.}
  \vspace{-1em}
  \label{tab:attention}
\end{table}

\noindent\textbf{Results on Unseen Environment} We additionally test our model's ability to generalize to unseen environments by evaluating our best model on a subset of data collected in a room with different acoustic and visual properties. The data follows the same structure as the main evaluation dataset, with 5 participants split into 2 conversation subgroups, and is $\sim$49 minutes. Further details are provided in the supplement. On this unseen environment, our best model achieves 80.43\% mAP, demonstrating scene generalizability.

\noindent\textbf{Spatial Modality Alignment in Self-Attention}
A key feature of our approach is leveraging multichannel audio to learn a spatial feature representation of voice activity in the scene, which we align with the visual representation by concatenating along the channel dimension. In our transformer design, therefore, each spatiotemporal patch of $\mathcal{Z}^{AV}$ contains the visual and audio features for that region. We validate this choice by designing an additional experiment where the modalities are not fused before the transformer, like in VisualBERT\cite{li2019visualbert}. 
Rather than concatenating $\mathcal{Z}^{A}$ and $\mathcal{Z}^{V}$ before flattening them into a sequence of tokens to pass to $\mathcal{T}$, we separately create a sequence of 182 tokens each from $\mathcal{Z}^{V}$ and $\mathcal{Z}^{A}$ and pass these 364 total tokens of dimension 512 to $\mathcal{T}$. 
We concatenate the modalities afterwards. In this case, the self-attention mechanism is left to discover alignment between the modalities itself by comparing all tokens across both modalities. We observe a decrease in performance, indicating that our choice to explicitly align the visual and audio modalities spatially contributes to our model's ability to use both modalities together successfully. This result suggests utility in incorporating inductive bias about the spatial alignment of visual and multichannel audio modalities into token design for transformer architectures.

\begin{figure}
  \centering
  \includegraphics[width=0.7\linewidth]{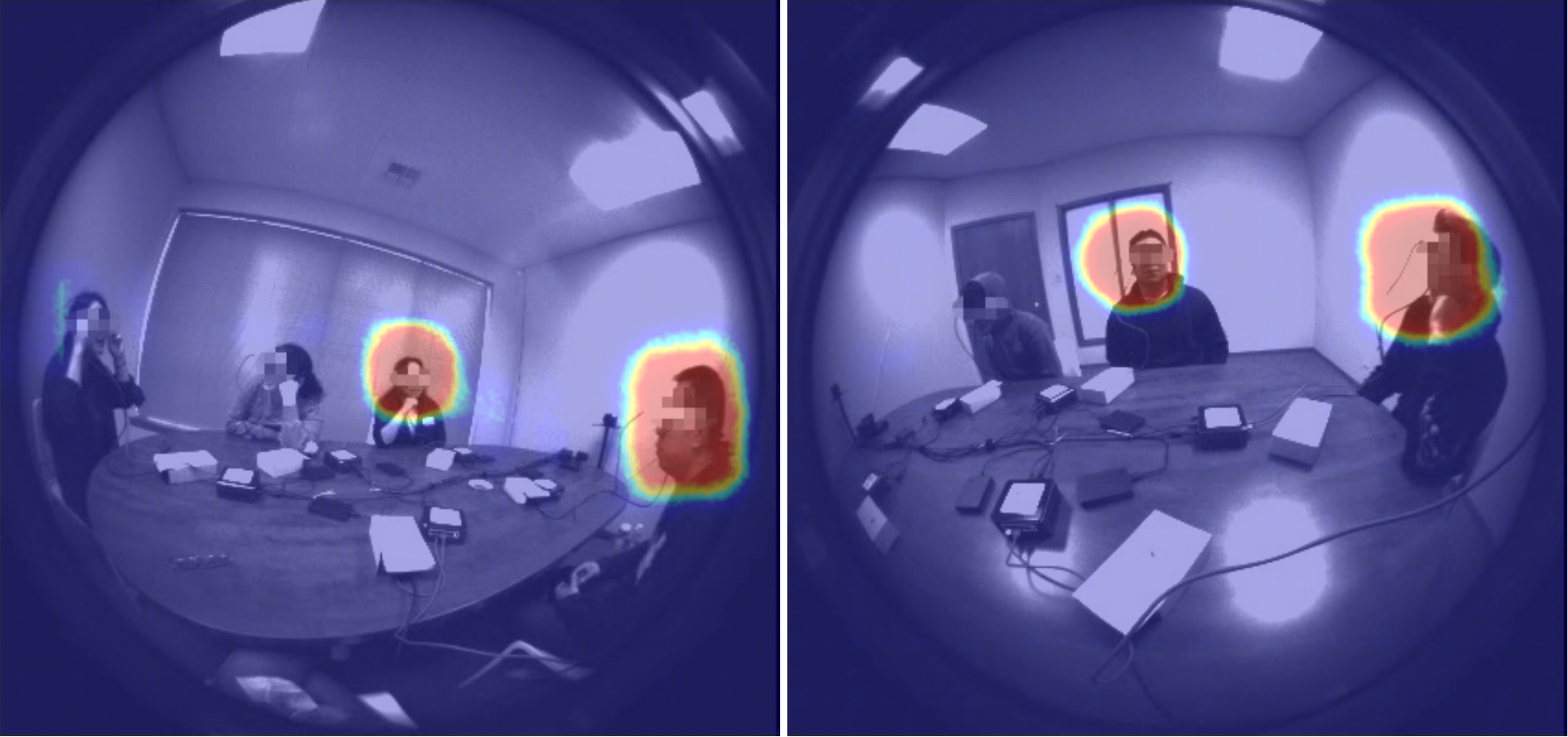}
  \caption{Decoupling SAAL \& ASL: Without the responsibility for learning ASL, our model learns to segment the people within the camera wearer's conversation group.}
  \vspace{-2em}
  \label{fig:perfectASL}
\end{figure}

\noindent\textbf{Decoupling SAAL \& ASL}
By nature of our dataset design, we train our model to predict auditory attention 
as the intersection of active speakers and people in the camera wearer's conversation group. To investigate the relationship between SAAL and ASL, we train a model by modulating our model's heatmap for the attended class with the ground truth ASL labels before calculating loss and performing backpropagation. The loss is calculated as
$ \mathcal{L}_{CE}(\mathcal{H}\textsubscript{pred} * \text{ASL}\textsubscript{label}, \mathcal{H}\textsubscript{label})$.
In this way we eliminate the responsibility for ASL from our model. The model achieves 92.67\% mAP on SAAL, however, this is a purely theoretical upper bound that relies on perfect ASL. We visualize $\mathcal{H}\textsubscript{pred}$ for our model before it is modulated by ASL\textsubscript{label} in Figure \ref{fig:perfectASL}. We observe that our model learns to more generally segment the camera wearer's conversation group regardless of speaking activity, and likely relies on \text{ASL}\textsubscript{label} to constrain the heatmaps to the exact bounding box region.

\section{Conclusion}
We introduce the novel problem of Selective Auditory Attention Localization (SAAL) and demonstrate that a deep model is able to perform this task on a challenging multi-speaker conversation dataset. We propose a multimodal modeling approach that uses egocentric video and multichannel audio to predict a person's target of auditory attention, combining spatial and temporal features from both modalities and using a transformer to reason holistically about the scene. We believe our work is an important step towards selective sound source enhancement and AR devices that can help people converse in noisy environments.

\noindent\textbf{Acknowledgments:} We thank Jacob Donley and Christi Miller for facilitating data collection, and Calvin Murdock and Ishwarya Ananthabhotla for valuable discussions.


{\small
\bibliographystyle{ieee_fullname}
\bibliography{egbib}
}

\clearpage

\setcounter{section}{0}
\renewcommand*{\thesection}{\Alph{section}}

\section*{Appendix}
In this appendix, we provide further details about elements of our work. We organize the content as follows:
\begin{itemize}
\itemsep0em
    \item \ref{sec:visualizations} - Visualizations
    \item \ref{sec:implementation} - Implementation Details
    \item \ref{sec:dataset} - Dataset Details
    \item \ref{sec:add_baselines} - Additional Baselines
    \item \ref{sec:realtime} - Real-Time Applications
    \item \ref{sec:societal} - Societal Impact
    \item \ref{sec:futurework} - Limitations \& Future Work
\end{itemize}

\section{Visualizations}\label{sec:visualizations}
Visualizations of our model's output heatmaps on videos in the test split of our dataset can be viewed on our project page \url{http://fkryan.github.io/saal}. These examples illustrate the complexity of the conversation environments in our dataset; we see that most frames contain multiple people within the FOV, and there are typically multiple people speaking at once. There is lots of head motion as people engage in head-nodding behaviors and look between different people and around the room while listening. Additionally, cases occur where multiple people within the camera wearer's conversation group speak at the same time and are both considered to be attended. Our model is able to determine the target(s) of attention effectively in many of these difficult cases and identifies temporal attention shifts between attended speakers.

These visualizations additionally give insight into failure modes. Our model has difficulty in some cases where attended and non-attended speakers are close together, and sometimes falsely identifies people as attention targets while the camera wearer is speaking and is not attending to any of the visible people. These failure modes reflect the challenging nature of our evaluation dataset and give insight into how future work may improve upon our approach.

\section{Implementation Details}\label{sec:implementation}
\subsection{Model}

\begin{figure}[h]
  \centering
   \includegraphics[width=1.0\linewidth]{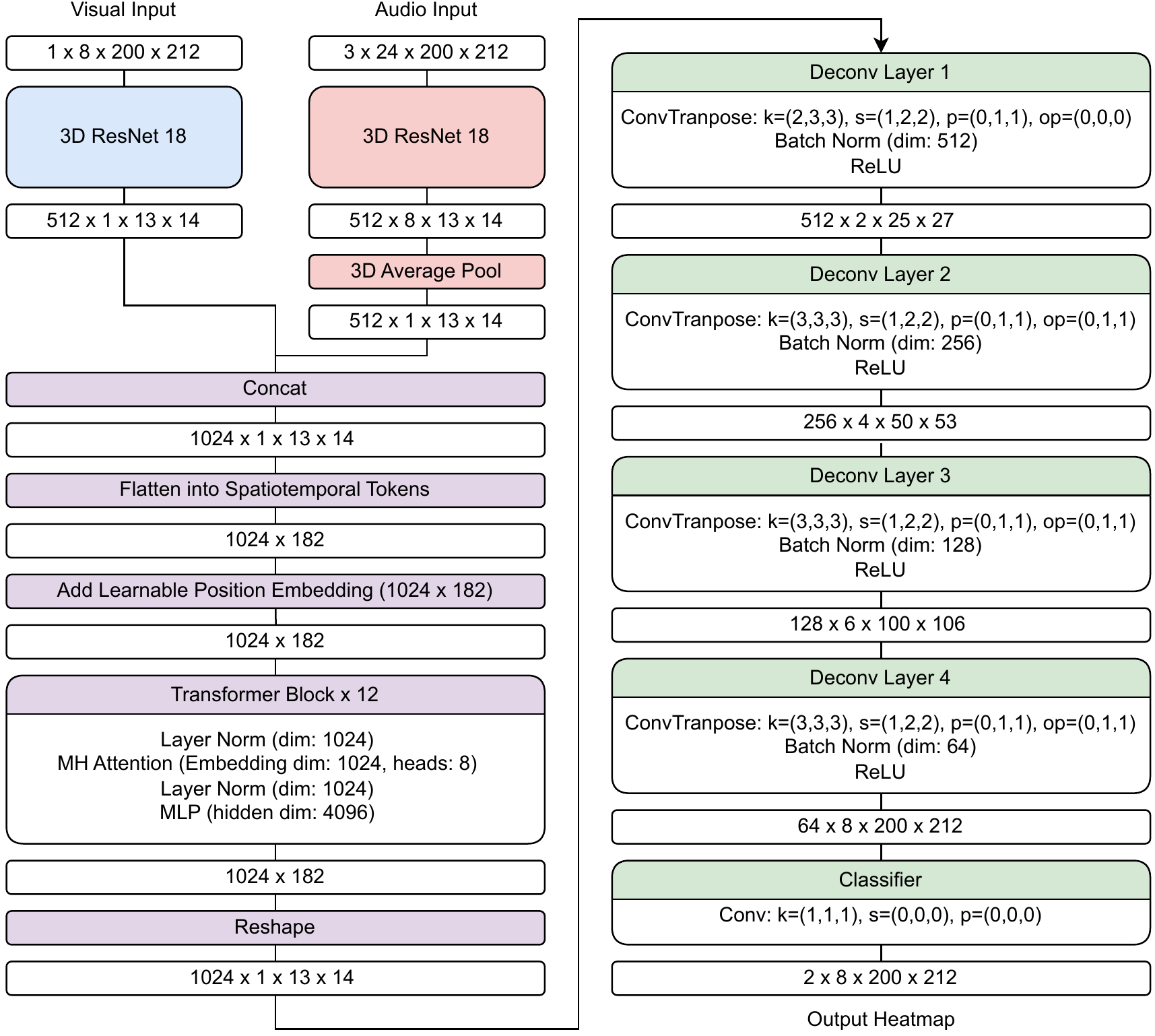}
   \caption{Architecture details. Shapes are denoted as channels $\times$ frames $\times$ height $\times$ width.}
   \label{fig:architecture_details}
\end{figure}

We provide architecture details for our model in Figure \ref{fig:architecture_details}, including the output shapes after each component and individual layer details. Our best model (\textbf{Heads} \& \textbf{Audio Corr + Spectogram}) converges after 5 epochs and takes approximately 1 hour per train epoch on 4 GPU's with batch size 32, using our input clip size of 8 visual frames with temporal stride 3 and 24 audio frames.

\subsection{Input Representations}

\begin{figure}[]
  \centering
   \includegraphics[width=1.0\linewidth]{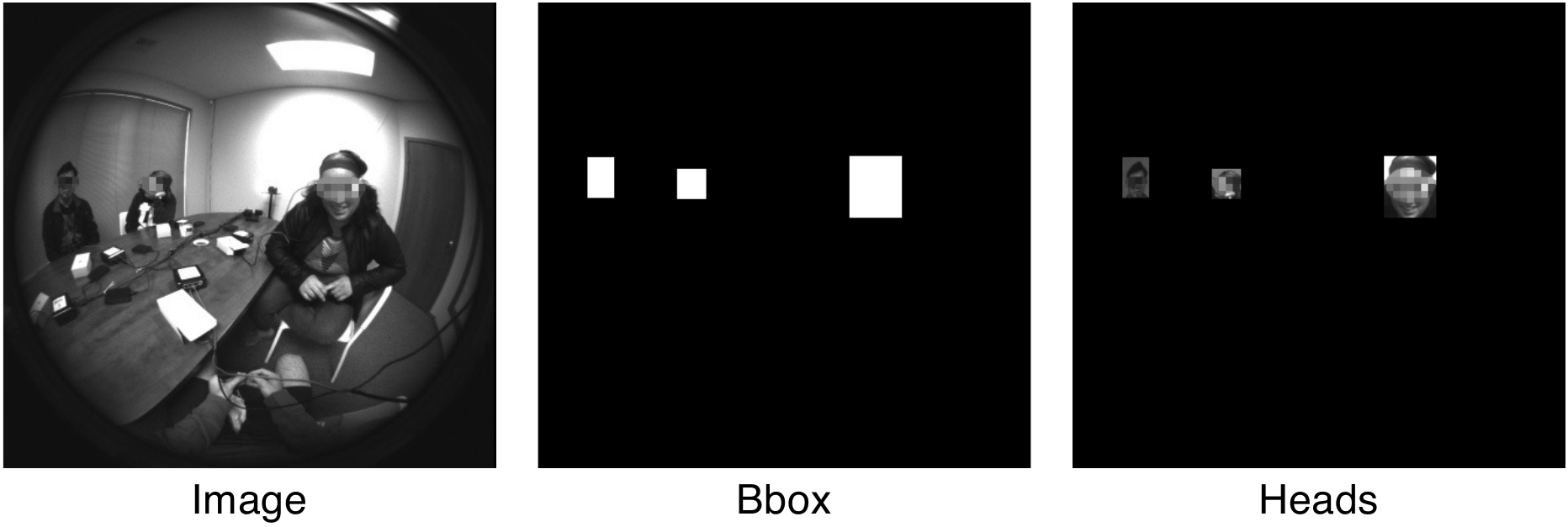}
   \caption{Visual input representations}
   \label{fig:visual_input}
\end{figure}

\noindent\textbf{Visual Input Representation}
In our experiments, we consider 3 visual input representations: \textbf{Image} (the raw image), \textbf{Bbox} (a binary map of the bounding boxes), and \textbf{Heads} (the cropped heads from the raw image on a black background). Visualizations of these 3 representations are shown for a sample frame in Figure \ref{fig:visual_input}.

\begin{figure}[]
  \centering
   \includegraphics[width=1.0\linewidth]{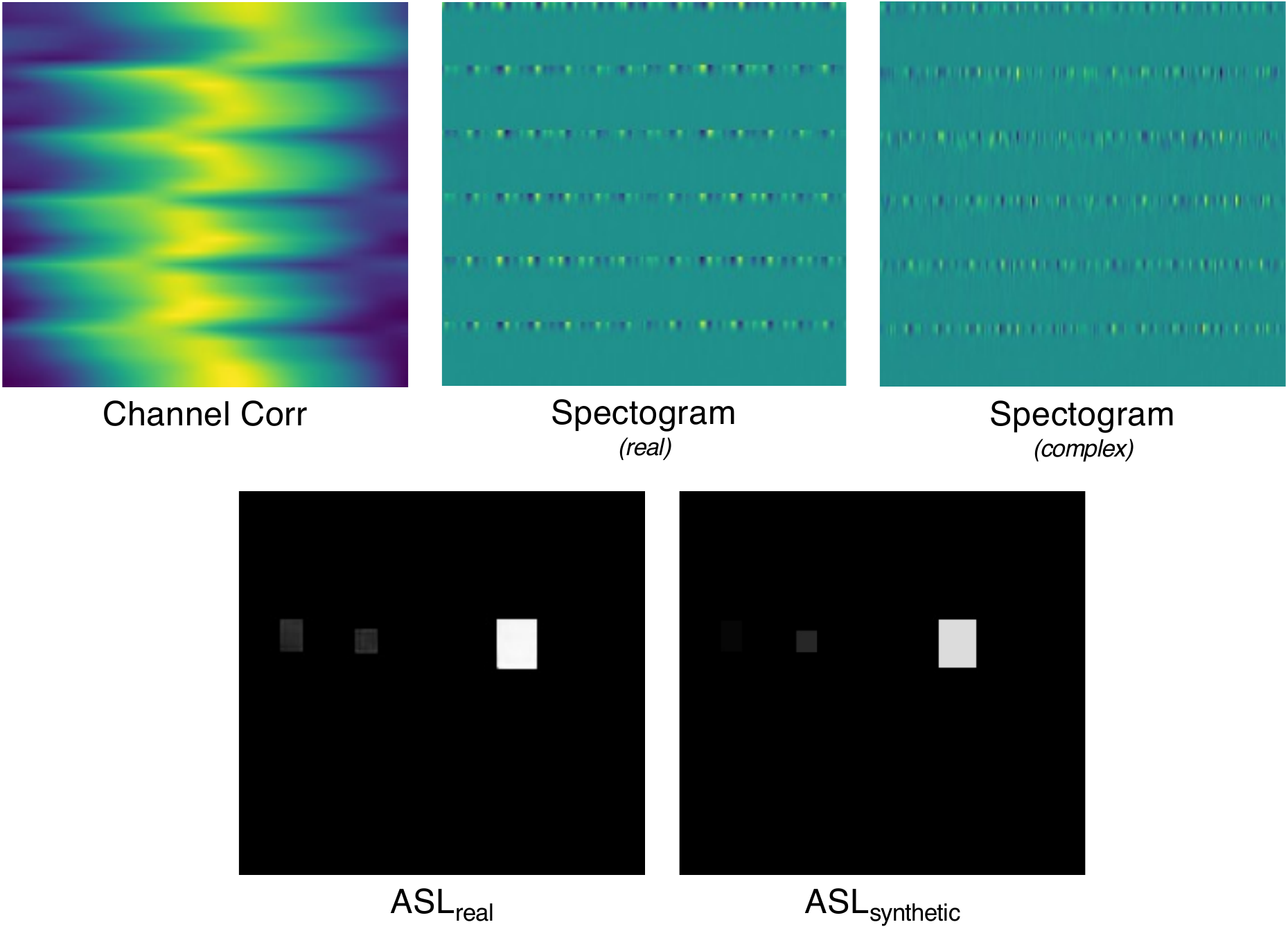}
   \caption{Audio input representations}
   \label{fig:audio_input}
\end{figure}

\noindent\textbf{Audio Input Representation}
In our experiments, we consider 4 audio input representations, where the representation for each audio frame corresponds to the 6-channel audio segment associated with an egocentric video frame. The input representations are \textbf{Channel Corr} (the channel correlation features), \textbf{Channel Corr + Spectogram} (the channel correlation features concatenated with the real and complex parts of the multichannel spectogram), \textbf{ASL\textsubscript{real}} (ASL output maps from MAVASL \cite{jiang2022egocentricdeep} trained on the the active speaker labels for our dataset), and \textbf{ASL\textsubscript{synthetic}} (MAVASL ASL maps trained on a synthetic dataset for our microphone array). We visualize these representations for a sample audio frame in Figure \ref{fig:audio_input}. The channel correlation features capture spatial audio information by representing the cross correlation between each pair of channels in the microphone array at each time. They are calculated in the same way as Jiang et al. \cite{jiang2022egocentricdeep}, which finds them to be an effective spatial audio input representation for ASL.

We augment these features with the real and complex parts of the spectogram to include finer grained details about the speech signals. To construct the multichannel spectograms we calculate the real and complex parts of the spectogram for each channel individually, $\mathcal{S}^{R}_{1}... \mathcal{S}^{R}_6$ and $\mathcal{S}^{C}_{1}... \mathcal{S}^{C}_6$, and concatenate these vertically to form the combined real spectogram $\mathcal{S}^{R}$ and complex spectogram ${S}^{C}$. We then concatenate $\mathcal{S}^{R}$ and ${S}^{C}$ along the channel dimension with the channel correlation features $\mathcal{C}$ to form the $3\times200\times212$ audio input feature for each frame. We additionally tried concatenating $\mathcal{S}^{R}_{1}... \mathcal{S}^{R}_6, \mathcal{S}^{C}_{1}... \mathcal{S}^{C}_6$ along the channel dimension instead of vertically along with $\mathcal{C}$ to form a $13\times200\times212$ input feature for each frame. We found this slightly reduced performance for our best model (81.68\% mAP as opposed to 82.94\% mAP).

\noindent\textbf{ASL\textsubscript{synthetic} Training} For the ASL map audio input representations, we include the \textbf{ASL\textsubscript{synthetic}} input representation in addition to \textbf{ASL\textsubscript{real}} to cover both the case where an ASL model is tuned to the dataset and when it is not. Because microphone array setups can vary widely between systems, an existing ASL model that uses multichannel audio may not be immediately applicable or available.

The synthetic multi-channel audio training data is generated using the VCTK speech dataset \cite{VTCKveaux2017cstr} and the far field 
ATFs (audio transfer functions) 
of the microphone array on the headset. We uniformly sample all the possible sound source directions and for each 
direction, we randomly 
select several speech samples from the VCTK dataset and apply the corresponding 
ATFs to generate the audio signal for each
microphone. This is equivalent to putting a virtual speech sound in each direction.
We use clean speech; we do not introduce noise and room acoustics effects.
The image data is generated using a cut-and-paste method. We paste a ``speaking''
head, randomly selected from the speaking heads in the EasyCom dataset \cite{EasyCom}, corresponding to the direction of the audio 
signal, on a black background. We also paste five more ``non-speaking'' heads from the EasyCom dataset in randomly selected positions
in the image. We randomize the size of the heads to simulate people at different distances from the wearer.
The ground truth 360-degree voice activity map and the ground truth of the voice activity map in the FOV can be easily generated using the known direction of the speech source and placement of the head. 
We use a total of 85,277 audio-visual training samples. The end-to-end training converges in 50 epochs with learning rate 1e-4 and the Adam optimizer.

\section{Dataset Details}\label{sec:dataset}
\subsection{Comparison to Prior Datasets}
To our knowledge, there is no existing dataset and accompanying labels that support our Selective Auditory Attention Localization task. EgoCom \cite{northcutt2020egocom} and EasyCom \cite{EasyCom} capture small group conversations with egocentric video and multichannel audio (binaural in the case of EgoCom) and include speech activity labels. However, both focus on single group conversations, with EgoCom containing conversations among 3 people and EasyCom containing conversations among groups of 3-6. In these single-group conversation scenarios, there are rarely cases of more than one person speaking nor are there speakers in the background (EasyCom does include background noise played through speakers, but there are not actual, visible people speaking in the background), so determining auditorily attended speakers can be effectively reduced to audiovisual Active Speaker Localization. However, this reduction is not suitable for realistic noisy conversation environments where there are multiple speakers present. Training a model for SAAL on these datasets therefore would not generalize to complex conversation environments, and evaluating a model for SAAL on these datasets cannot reflect the model's ability to identify selective auditory attention among competing speakers. However, competing speaker environments like restaurants and large group social settings are a main target for downstream sound source enhancement applications. In this work, we specifically seek to investigate selective auditory attention in the presence of multiple speakers, where a person must selectively attend to certain speaker(s) and tune out others. We therefore choose to collect and evaluate on a dataset that explicitly captures overlapping speech, multiple simultaneous conversations, and visible background speakers, where ASL alone cannot determine auditory attention.

The AV Diarization \& Social Interactions benchmark subset of Ego4D \cite{ego4dshort} contains a broader array of conversation scenarios captured by egocentric video and, in limited cases, binaural audio. The dataset includes some cases with multiple speakers and background speakers such as grocery stores, outdoor dining areas, office hours, and group board games. However, determining auditory attention labels for such a dataset is subjective and ambiguous, and there are relatively few cases that capture multiple conversations occurring at once. Quantitatively, only 1.9\% (65,261 frames) of the Ego4D train dataset contains at least 2 visible speakers. Ego4D does include ``Talking to me" labels which identify which speaker(s) are talking to the camera-wearer and may give insight into who the camera wearer is interacting with and who is in the background. However, this label is inherently different than auditory attention, which identifies listening behavior. Within the Ego4D train dataset, only 0.24\% (8,444) frames have a visible ``talking to me" speaker as well as another visible speaker. While the``talking to me" labels could indicate which speakers are in a conversation group with the camera-wearer, it is clear there are few cases that capture competing speakers where we could potentially determine from the labels which speaker the camera wearer is listening to. Additionally, the Ego4D dataset largely does not contain multichannel audio, which we find to be a critical aspect of our modeling approach. We evaluate our approach on a dataset we design to capture multi-conversation scenarios on a large scale with selective auditory attention labels.

\subsection{Conversation Layouts}\label{subsec:layouts}

\begin{figure}[h]
  \centering
   \includegraphics[width=1.0\linewidth]{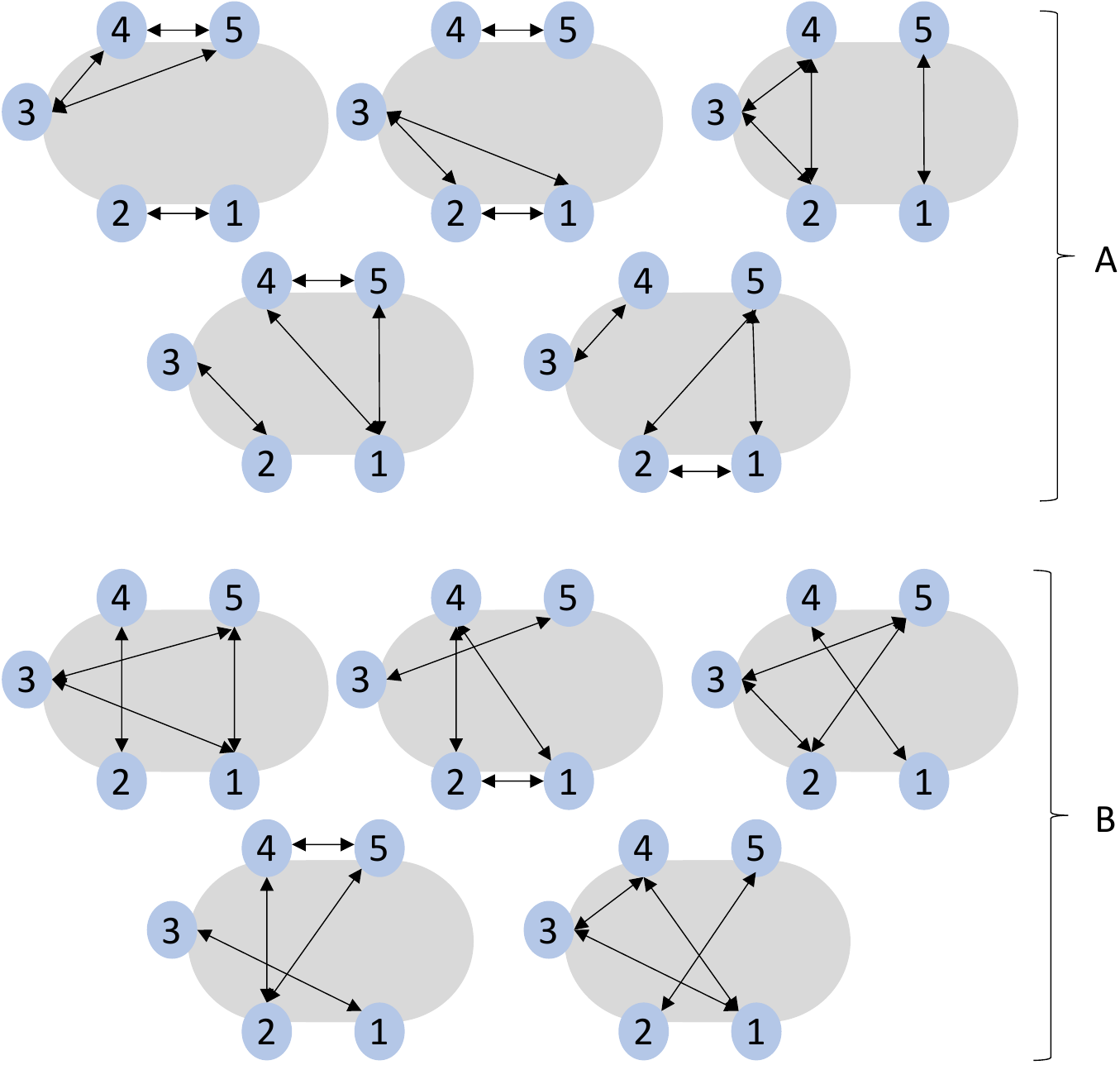}
   \caption{Conversation layouts: Group A represents common scenarios where people converse only with those adjacent or across to them, while Group B represents more challenging cross-talk scenarios. Group A comprises approximately 2/3 of our dataset.}
   \label{fig:layouts}
\end{figure}

We train and evaluate on a dataset of 5-person scenarios where people converse in 2 separate conversation groups. We illustrate the different conversation group assignments we use in Figure \ref{fig:layouts}. We design the dataset to encompass two kinds of conversational scenarios, which are denoted as Group A and Group B. The conversation layouts in Group A reflect situations where separate social groups converse in close proximity, such as at a coffee shop. In these settings, there is spatial separation between the 2 groups and people converse with those adjacent or directly across from them. The conversation layouts in Group B represent more challenging scenarios where the people in the different conversation groups must talk across each other. These are representative of situations like a large dinner table where multiple conversations occur simultaneously. Group A layouts comprise approximately 2/3 of our dataset and Group B layouts comprise approximately 1/3. In this way, the majority of our dataset is Group A layouts, which are simpler and more likely occur, but the dataset also encompasses challenging cases like those in Group B.

\subsection{Person Tracking, Head Bounding Box, \& Active Speaker Labels}

To label our auditory attention dataset, we need to track each person throughout the video.
We take advantage of the camera pose output from the Intel SLAM camera for robust tracking.
If people are perfectly still and the wearer only rotates their head, we can track each person's
location by back-projecting to 3D, rotating, and then projecting to the 2D image. In reality, 
this does not work because people move and the wearer's head not only rotates but also translates.
Our tracking-by-detection method back-projects head detection to the surface of a 2-meter sphere.
The head bound box detector is Yolo-v3-tiny \cite{adarsh2020yolo} trained on images from the Open Images dataset \cite{kuznetsova2020open}. 
By using such a representation, the tracking algorithm generates a matching cost matrix 
using the current object observations and previous position estimations. Using min-cost matching,
we extend each target's trajectory frame-by-frame. 

We use the wearer voice activity classification network from \cite{jiang2022egocentricdeep} to determine voice activity for each person, giving us the set of active speakers at each time. Because the headset microphones can easily pick up the sound of the camera wearer speaking, we find this to be a reliable method for determining active speakers. The model is trained synthetically for our headset's microphone array using speech data from VCTK. The near-microphone ATFs are used to generate positive training examples and the far-field ATFs for negative training examples. We manually inspected the tracking, bounding box, and speaker labels to ensure high quality.

\subsection{Unseen Environment Data Subset} To test the generalizability of our model under different visual and acoustic conditions, we collected a small subset of data in a different room, which resembled an open kitchen area. The structure of this subset followed that of the main dataset, with 5 participants conversing in 2 simultaneous conversation subgroups. The participants are different than those included in the main dataset. Conversation layouts from both Group A layouts and Group B layouts were included (see Section \ref{subsec:layouts}), and ground truth labels were constructed in the same manner as in the main dataset. However, only the 3 people in one of the conversation subgroups wore headsets. In total, this subset included 87,977 frames, or ~49 minutes of data. The result of 80.43\% mAP on this dataset was obtained by running our best model trained on the main dataset on the unseen environment subset with no finetuning. This result shows that our model can generalize to a different environment as well as to cases where people are not wearing glasses.

\begin{table}
\small
  \centering
  \begin{tabular}{@{}lc@{}}
    \toprule
    Method & mAP (\%) \\
    \midrule  
    Perfect ASL*      & 47.99 \\
    CP--I*  & 63.55  \\
    CP--II* & 51.48  \\
    CS--I* & 53.86 \\
    CS--II* & 49.47 \\
    \hline
    LS--I* & 27.78 \\
    LS--II* & 30.63 \\
    \hline
    FCN+ASL\textsubscript{real} + CP      & 72.33 \\
    FCN+ASL\textsubscript{synthetic} + CP & 74.30 \\
    FCN+ASL\textsubscript{synthetic} + CP + WVA* & 73.32 \\
    \hline
    MAVASL--I         & 59.11  \\
    MAVASL--II        & 75.20  \\
    MAVASL--III       & 72.90   \\
    \hline
    Ours--Bbox \& ASL\textsubscript{synthetic} & 75.93 \\
    Ours--Bbox \& ASL\textsubscript{real} & 74.97 \\
    Ours--Bbox \& Channel Corr & 80.41 \\
    Ours--Bbox \& Channel Corr + Spectogram & 80.31 \\
    Ours--Image \& ASL\textsubscript{synthetic} & 72.20 \\
    Ours--Image \& ASL\textsubscript{real} & 70.04 \\
    Ours--Image \& Channel Corr & 76.52 \\
    Ours--Image \& Channel Corr + Spectogram & 76.95 \\
    Ours--Heads \& ASL\textsubscript{synthetic} & 76.72 \\
    Ours--Heads \& ASL\textsubscript{real} & 77.11 \\
    Ours--Heads \& Channel Corr & 82.35 \\
    \textbf{Ours--Heads \& Channel Corr + Spectogram} & \textbf{82.94} \\
    \bottomrule
  \end{tabular}
  \caption{Comparison results for all methods on the multi-speaker conversation dataset. (*) denotes methods that use inputs that are not given to our model including ground truth active speaker labels, the camera wearer's speaker activity label, and other people's headset audio.}
  \label{tab:full_comparison}
\end{table}

\section{Additional Baselines}\label{sec:add_baselines}
In addition to the baselines described in section 4.2 we provide two further groups of baselines. A full comparison of all competing methods, including those described in the main paper, is shown in Table \ref{tab:full_comparison}.

\textbf{(1)} Fully convolutional network (FCN) combined with ASL (FCN+ASL): We investigate the extent to which a simple convolutional architecture can solve our task with different types of inputs by adapting the FCN AV Network architecture from MAVASL\cite{jiang2022egocentricdeep} to predict auditory attention from a concatenation of the raw image and a pre-predicted ASL map (either \textbf{ASL\textsubscript{real}} or \textbf{ASL\textsubscript{synthetic}}), and include an additional center distance map channel to embed the center in which each pixel's value equals the normalized distance to the center of the image, denoted as (CP). We additionally include a variation where the wearer's ground truth speech activity is represented as an extra input channel in which each pixel is 0 or 1 depending on wearer's voice activity label (WVA), which is the label for whether the wearer is speaking or not.

\textbf{(2)} Selecting attended speaker based on close-microphone speech activity: We additionally estimate the loudness of each person relative to the camera wearer's position by using the audio energy of each person's worn microphones on their headset and their distance from the camera wearer, as calculated by the SLAM camera. We calculate loudness for each visible person as $L = \frac{A}{d^2}$ where $A$ is the short-time energy from the person's wearable microphone array (averaged across the channels) for the given audio frame, and $d$ is the distance between the person and the camera-wearer, as estimated by the SLAM camera. We construct 2 baselines that use this loudness measure: LS--I selects the loudest speaker as attended. LS--II selects the loudest speaker as attended unless the wearer is speaking per the ground truth voice activity labels, in which case it selects no people as attended. We note that this baseline uses inputs not given to our model: the audio signals from the headsets of the other participants, the distance as calculated by the SLAM camera, and the ground truth voice activity label for the camera wearer. It is thus not a fair comparison to our model, but we include it to illustrate that in complex conversation environments, assuming that the loudest speaker is attended is not sufficient to solve SAAL.

\section{Real-time Applications}\label{sec:realtime}
Our problem is motivated by the application of selective sound source enhancement, or developing devices that can enhance certain sound sources while suppressing others. Such a setting demands algorithms that can be run in real-time. While we do not implement our architecture specifically to run in real-time on a mobile device in the scope of this work, our problem formulation reflects this downstream application in two ways: (1) In contrast to the AVA-Active Speaker detection problem formulation which classifies a single head bounding box track at a time, our model reasons about the full scene and all people at once. Not only does this modeling choice reflect the need to reason holistically about the scene to determine SAAL, but this is also conducive to efficient real-time applications. (2) Our architecture is a clip-based video model that runs on a short temporal window at a time (approximately 1 second). Our architecture can be adapted to produce frame-level predictions using this short temporal history.  Future work may explore implementing our architecture as part of a real time sound source enhancement system.

\section{Societal Impact}\label{sec:societal}
Our work is motivated by developing wearable computing devices that can help people communicate naturally in noisy environments, and can especially assist individuals with hearing difficulties in day to day conversations. 
Researchers have explored hearing enhancement systems that allow a user to select certain sound sources to enhance using controls like head orientation, eye gaze, and hand controls \cite{favre2018improving, hart2009attentive, best2017benefit, kidd2017enhancing, kidd2013design, ricketts1999comparison, hladek2019interaction, geronazzo2020superhuman}. 
By using egocentric cues to automatically determine attended speakers as sound sources to enhance, our work may allow more naturalistic behavior while using such a system. 
We acknowledge that selective sound source enhancement brings about important privacy considerations. 
The ability to enhance certain sounds may change notions of conversational privacy in public places, and care must be taken in implementing devices with such capabilities. 
We note that our work does not apply our algorithm to an end-to-end selective sound source enhancement system.

Additionally, in our dataset design, we avoid constructing scenarios such as intentional, covert eavesdropping that could be used to implement systems with the intent to violate privacy. 
We instead focus on modeling listening behaviors in multi-group conversation scenarios where participants expect to be heard by others. 
We believe our work has great potential for the development of devices that assist people with everyday social communication, and can especially help individuals with hearing loss.

\section{Limitations \& Future Work}\label{sec:futurework}
While our work takes an important step towards selective sound source enhancement by addressing modeling selective auditory attention from an egocentric audiovisual perspective for the first time, our approach has limitations. In our dataset design, we constrain our ground truth auditory attention labels to considering speakers within the camera wearer's conversation group as being attended. In reality, this does not encompass attentional dynamics like getting distracted by speakers in the other conversation or sounds occurring elsewhere in the room (which falls under bottom-up auditory attention), or simply zoning out of the conversation. Because auditory attention is covert, sourcing true labels for auditory attention that encompass all such cases is impossible. We believe we take a practical approach to generating objective labels for selective auditory attention, and our labels are appropriate for downstream selective sound source enhancement applications in conversational settings.

There are several opportunities for future work in this direction to improve upon our approach. First, our method may be applied to settings beyond what we collect in our dataset including larger group settings, diverse physical environments, conversing while doing other activities, changing conversation groups over time, and scenarios where people move around the space. Additionally, we constrain our attention modeling to in-FOV cases. While this works well given the wide 180-degree FOV camera we use, future work may explore expanding our model's capabilities to handle cases where people move beyond the wearer's FOV. Further work may also explore explicitly modeling longer-term context, such as conversational turn taking and social groupings over time, to improve predictions. We hope our work will inspire further research in this exciting direction.

\end{document}